\documentclass[a4paper,fleqn]{cas-dc}

\usepackage[authoryear,longnamesfirst]{natbib}
\usepackage{graphicx}
\usepackage{subcaption}
\usepackage{amsmath}
\usepackage{pifont}
\usepackage{bm}
\usepackage{silence}

\def\tsc#1{\csdef{#1}{\textsc{\lowercase{#1}}\xspace}}
\tsc{WGM}
\tsc{QE}
\tsc{EP}
\tsc{PMS}
\tsc{BEC}
\tsc{DE}
\usepackage{orcidlink}



\begin{document}
\let\WriteBookmarks\relax
\def\floatpagepagefraction{1}
\def\textpagefraction{.001}



\title [mode = title]{
Changes in Gaza: DINOv3-Powered Multi-Class Change Detection for Damage Assessment in Conflict Zones
}





\newcommand{\ORCIDsup}[1]{\texorpdfstring{\textsuperscript{\orcidlink{#1}}}{}}
\author[a]{Kai Zheng}
\fnmark[1]
\author[b]{Zhenkai Wu}
\fnmark[1]
\author[c]{Fupeng Wei\ORCIDsup{0000-0002-2337-1483}}
\fnmark[1]
\author[d]{Miaolan Zhou}
\author[e]{Kai Li}
\author[f]{Haitao Guo}
\author[f]{Lei Ding}
\author[a,b]{Wei Zhang}
\cormark[1]

\author[g,h]{Hang-Cheng Dong\ORCIDsup{0000-0002-4880-6762}}
\cormark[1]

\affiliation[a]{organization={School of Computer Science and Technology},            organization={Zhejiang University}, 
    city={Hangzhou},
    postcode={310027}, 
    country={China}}


\affiliation[b]{organization={School of Software Technology},            organization={Zhejiang University}, 
    city={Hangzhou},
    postcode={310027}, 
    country={China}}


\affiliation[c]{organization={School of Information Engineering},            organization={North China University of Water Resources and Electric Power}, 
    city={Zhengzhou},
    postcode={450046}, 
    country={China}}


\affiliation[d]{organization={Polytechnic Institute},            organization={Zhejiang University}, 
    city={Hangzhou},
    postcode={310027}, 
    country={China}}


\affiliation[e]{organization={Institute of Systems Engineering},            organization={Academy of Military Sciences}, 
    city={Beijing},
    postcode={100171}, 
    country={China}}


\affiliation[f]{organization={Department of Geo-spatial Information},            organization={Information Engineering University}, 
    city={Zhengzhou},
    postcode={450001}, 
    country={China}}


\affiliation[g]{organization={School of Instrumentation Science and Engineering},            
 organization={Harbin Institute of Technology}, 
    city={Harbin},
    postcode={150001}, 
    country={China}}


\affiliation[h]{organization={Harbin Institute of Technology Suzhou Research Institute},            
    city={Suzhou},
    postcode={215100}, 
    country={China}}

\cortext[cor1]{Wei Zhang and Hang-Cheng Dong are co-corresponding authors.}

\fntext[fn1]{Kai Zheng, Zhenkai Wu, and Fupeng Wei are co-first authors.}


\begin{abstract}
Accurately and swiftly assessing damage from conflicts is crucial for humanitarian aid and regional stability. In conflict zones, damaged zones often share similar architectural styles, with damage typically covering small areas and exhibiting blurred boundaries. These characteristics lead to limited data, annotation difficulties, and significant recognition challenges, including high intra-class similarity and ambiguous semantic changes. To address these issues, we introduce a pre-trained DINOv3 model and propose a multi-scale cross-attention difference siamese network (MC-DiSNet). The powerful visual representation capability of the DINOv3 backbone enables robust and rich feature extraction from bi-temporal remote sensing images. The multi-scale cross-attention mechanism allows for precise localization of subtle semantic changes, while the difference siamese structure enhances inter-class feature discrimination, enabling fine-grained semantic change detection. Furthermore, a simple yet powerful lightweight decoder is designed to generate clear detection maps while maintaining high efficiency. We also release a new Gaza-change dataset containing high-resolution satellite image pairs from 2023-2024 with pixel-level semantic change annotations. It is worth emphasizing that our annotations only include semantic pixels of changed areas. Unlike conventional semantic change detection (SCD), our approach eliminates the need for large-scale semantic annotations of bi-temporal images, instead focusing directly on the changed regions, which terms multi-class change detection (MCD). We evaluated our method on the Gaza-Change and two classical datasets: SECOND and Landsat-SCD datasets. Experimental results demonstrate that our proposed approach effectively addresses the MCD task, and its outstanding performance paves the way for practical applications in rapid damage assessment across conflict zones.


\end{abstract}




\maketitle


\section{Introduction}

Accurately and timely assessing damage zones in conflict areas is a critical task with profound implications for humanitarian assistance, disaster relief, and post-conflict reconstruction~\cite{QING2022102899, HOLAIL202548}. Similar to building damage assessment in natural disasters~\cite{han2025multi}, remote sensing images, particularly high-resolution satellite data, have become an indispensable tool for large-scale monitoring of these changes. However, whereas previous building damage assessments caused by natural disasters focused more on binary changes, damage assessment in conflict areas may place greater emphasis on fine-grained types. Therefore, the core task in conflict zones is semantic change detection (SCD) of buildings.


In recent years, the remarkable success of deep learning in both computer vision~\cite{od2022review, sultana2020evolution} and natural language processing~\cite{guo2025deepseek} has profoundly reshaped the landscape of remote sensing semantic change detection. The evolution of SCD methodologies has largely mirrored broader trends in visual recognition. The field initially adopted convolutional neural network (CNN)-based architectures, drawing direct inspiration from semantic segmentation tasks~\cite{peng2019end}. However, these early, multi-stage pipelines often suffered from significant error accumulation. This limitation prompted a shift towards siamese network structures, which emerged as a more robust foundational paradigm for direct change representation~\cite{DLCD2018FC}. Subsequently, with the rise of vision transformers (ViTs)~\cite{vaswani2017attention, han2022survey, liu2021swin}, transformer-based architectures have begun to establish a new state-of-the-art for SCD. Nevertheless, a fundamental challenge persists: these data-hungry deep models require vast amounts of meticulously annotated data, which is notoriously difficult and expensive to obtain for remote sensing applications.


To address the aforementioned challenges, we introduce the multi-class change detection (MSD). Distinct from conventional semantic change detection (SCD), MCD eliminates the need for annotating entire semantic regions, instead focusing solely on change masks. This framework represents a direct extension of binary change detection (BCD). While significantly reducing annotation difficulty and time requirements, this new paradigm consequently increases the challenge of limited target region proportions. This necessitates substantial improvements in the model's capability to extract features from small target areas.

\begin{figure*}[ht]
  \centering
  \centering
  \includegraphics[height=4cm]{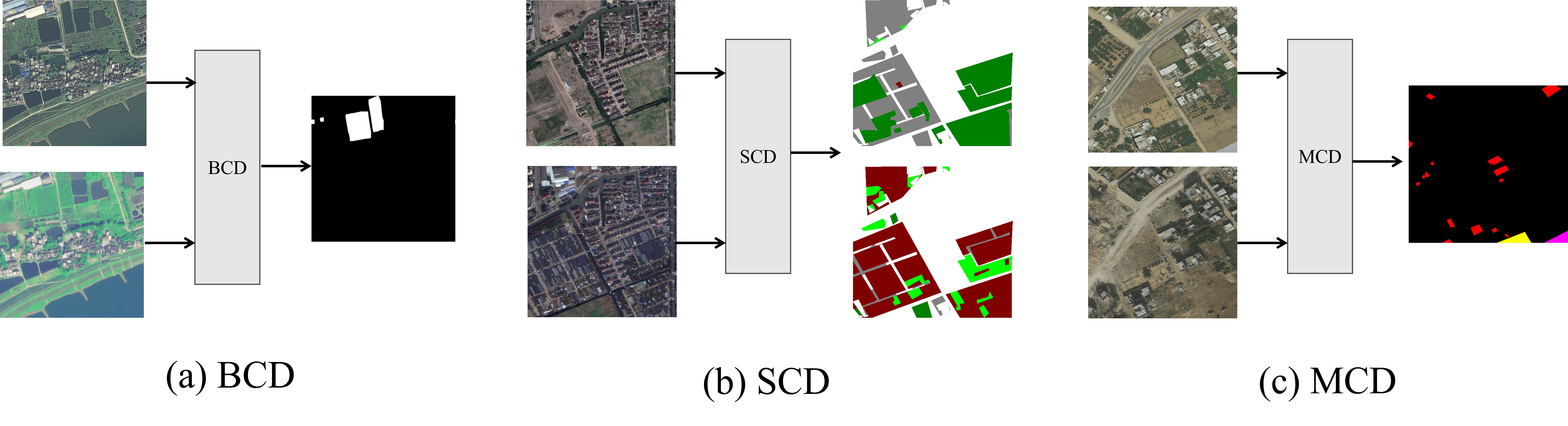}
  \caption{Evolution of change detection paradigms: (a) Binary Change Detection (BCD), (b) Semantic Change Detection (SCD), and (c) Multi-Class Change Detection (MCD).}
  \label{fig:csd}
\end{figure*}


Based on the above analysis, we identify four major challenges in framing conflict-induced damage assessment as a MCD task: (1) Inherent data scarcity: Limited by the geographical extent of conflict zones and the number of destroyed areas available for training. (2) Small target regions: MCD focuses exclusively on damaged areas, resulting in minimal semantic region coverage. (3) Subtle and ambiguous changes: Infrastructure damage in conflict zones varies significantly in severity and extent, particularly making minor damage difficult to detect. (4) High inter-class similarity: Different facility categories within the same region may share similar characteristics, making fine-grained damage assessment particularly challenging for semantic change detection.

\begin{figure*}[ht]
  \centering
  \begin{minipage}{1.0\linewidth}
  \centering
  \includegraphics[height=12cm]{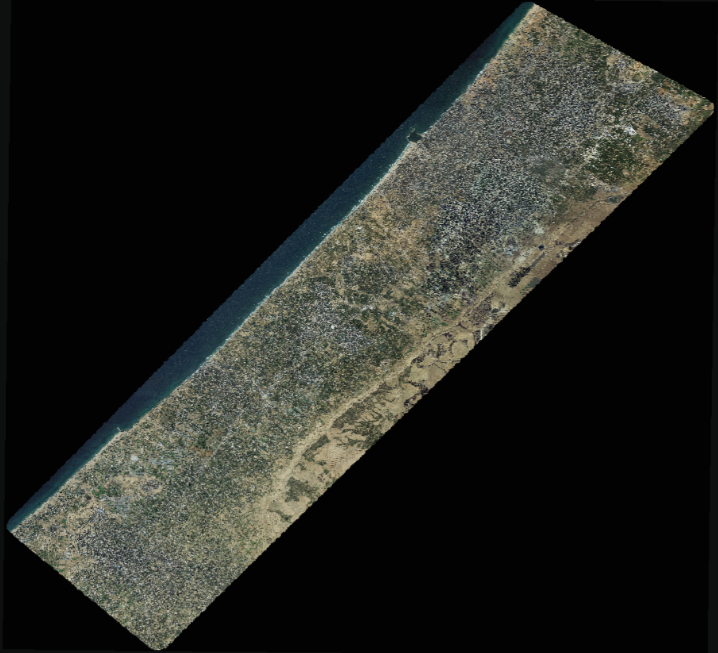}
  \end{minipage}\\[8pt]
  \caption{The panoramic remote sensing satellite image of the Gaza Strip.}
  \label{fig1}
\end{figure*}

To bridge this gap, we draw inspiration from the recent success of foundational vision models. We argue that leveraging their rich, pre-trained representations is key to overcoming data scarcity and recognizing subtle semantic changes. In this paper, we propose a novel DINOv3-driven siamese network for MCD. Specifically, we adopt the DINOv3~\cite{simeoni2025dinov3} model pre-trained on satellite data, with ConvNeXt~\cite{liu2022convnet} as its main backbone architecture, which helps reduce the distribution discrepancy between the pre-training data and the actual application data. Then, we propose a multi-scale attention mechanism to extract and enhance features at different levels, aiming to capture the subtle and ambiguous change features of infrastructure damage. Furthermore, we perform an absolute value differential operation on the obtained semantic-rich feature maps to increase inter-class feature differences. Finally, a carefully designed decoder network with attention enhancement is used to generate clear semantic change detection maps. We also release a building semantic change detection dataset of the Gaza area from 2023 to 2024. As shown in Figure \ref{fig1}, we present panoramic remote sensing images of the Gaza Strip captured by satellites. To the best of our knowledge, this is the first remote sensing semantic change detection study focused on conflict area assessment, laying a foundation for future research in related fields. In summary, our work makes the following key contributions:

\begin{itemize}

\item We introduce a multi-scale cross-attention difference siamese network (MC-DiSNnet). Built upon a pre-trained DINOv3 backbone, our network extracts robust, generalized features. The cross-attention mechanism is strategically employed to fuse multi-scale temporal features, enabling it to pinpoint subtle, semantic-changing regions effectively.

\item We contribute a new dataset for the Gaza area, containing high-resolution bi-temporal satellite image pairs from 2023-2024 with meticulously annotated pixel-level semantic change labels. To our knowledge, this is the first change detection study specifically focused on conflict area assessment.

\item We introduce the multi-class change detection (MCD) paradigm for damage assessment that fundamentally shifts from exhaustive bi-temporal semantic annotation to focused labeling of changed semantic regions. This strategic simplification significantly reduces annotation complexity and human labor. 

\item Extensive experiments show that our method achieves state-of-the-art performance not only on our proposed Gaza dataset but also on the public benchmarks, SECOND~\cite{yang2021asymmetric} and Landsat-SCD~\cite{yuan2022transformer}, demonstrating its superior robustness and generalization capability.
\end{itemize}

\section{Related Work}

In this section, topics related to remote sensing image change detection (RSCD) are introduced, including methods based on classical AI approaches and those based on vision foundation models.

\subsection{Classical Deep Learning for Change Detection}

Change detection involves multiple inputs. Traditional methods typically require multiple processing stages. To avoid error accumulation effects, \cite{peng2019end} pioneered the use of segmentation networks for change detection tasks. By pairing the input images, this single-stage approach not only shortens the detection pipeline but also enhances detection performance. Subsequently, models based on classical convolutions and Transformer models focusing on global information have emerged as key research foci in change detection. \cite{DLCD2018FC} employed a siamese network for change detection, as its architecture is inherently suited to this task. On the other hand, \cite{DLCD2021transformer} introduced a Transformer module after the CNN features to handle global information, addressing the challenges posed by high-resolution remote sensing images. WNet~\cite{tang2023wnet} builds upon this CNN-Transformer hybrid architecture by incorporating deformable convolutions and designing a feature fusion module (CTFM) to integrate local, global, and cross-scale features from CNN and Transformer encoders.

To achieve better fusion within such hybrid models, \cite{feng2022icif} proposed a parallel architecture integrating CNN and Transformer to simultaneously capture local and global features. Specifically, an intra-scale cross-interaction module was designed to interact with convolutional and Transformer features, followed by an inter-scale feature fusion module for integration. In contrast, \cite{he2024cbsasnet} devised a channel-bias separation attention (CBSA) module, which enhances the extraction of detailed information by integrating features from multiple receptive fields. \cite{changeformer} introduced ChangeFormer, a novel architecture combining Transformer with multi-layer perceptron (MLP). This model effectively captures both spatial and temporal features, making it well-suited for remote sensing change detection tasks. Focusing on urban scenarios, \cite{zhan2025difference} augmented multi-scale feature perception capabilities to detect urban building features of varying scales in remote sensing imagery.

\subsection{Vision Foundation Models}




Visual Foundation Models (VFMs) \cite{VFM} represent a category of AI models built upon deep neural networks and self-supervised learning. Trained on large-scale, multi-source datasets, they excel at capturing complex visual features and demonstrate superior performance across diverse visual tasks, including object detection, image classification, and semantic segmentation. Through pre-training on vast datasets, VFMs acquire general recognition capabilities that readily transfer to various downstream tasks. Beyond achieving state-of-the-art results in standard supervised learning, VFMs also facilitate advanced functionalities like zero-shot and interactive segmentation\cite{samclip}.

The advent of VFMs has significantly transformed deep learning-based visual processing. Masked Autoencoders (MAE) pioneered a vision masking strategy utilizing Transformers, achieving outstanding performance by heavily masking large portions of input images. CLIP \cite{clip} leverages massive image-text pairs for training, effectively boosting image-text similarity through metric learning. Segment Anything Model (SAM) \cite{sam} and its derivatives (e.g., FastSAM \cite{fastsam}, MobileSAM \cite{mobilesamv2}) mark a paradigm shift in image segmentation, demonstrating remarkable proficiency in detecting, segmenting, and generating objects of any category, showcasing exceptional versatility.


For remote sensing scenarios, visual foundation models (VFMs) still require effective fine-tuning to achieve satisfactory performance. Pioneering the application of VFMs to remote sensing change detection, \cite{SAMCD} proposed the SAM-CD architecture. This approach employs a lightweight version of FastSAM as the visual encoder and incorporates a convolutional adaptor to enhance its adaptability to remote sensing scenes. Building upon FastSAM, \cite{gao2025combining} utilized fine-tuning adapters to reduce data dependency. Conversely, \cite{zhang2024integrating} introduced DoRA (weight-decomposed low-rank adapter) for fine-tuning and designed a boundary distance-based loss function to optimize segmentation boundaries, specifically addressing the challenges of small objects and ambiguous edges. Focusing on lightweight performance, \cite{mobilesambased1} and \cite{SCDSAM} developed efficient yet effective models based on MobileSAM.

\section{Method}

\subsection{Overall Architecture}

The proposed multi-scale cross-attention difference sia-
mese network (MC-DiSNet), based on the DINOv3~\cite{simeoni2025dinov3} backbone, employs a two-branch encoder-decoder architecture consisting of three key components: an enhanced DINOv3 backbone, a multi-scale cross-attention difference (MSCAD) module, and a lightweight yet powerful decoder. The enhanced backbone preserves three hierarchical stages and injects change awareness using bottleneck adapters, multi-stage prompts, and LoRA.

The MSCAD module processes dual-branch features from the encoder. It aligns the tri-stage features with a cross-scale adapter, employs a diff-processor to capture temporal discrepancies, and integrates direct and adaptive differences via a diff-aggregator to produce sharp multi-scale change cues.

Subsequently, the decoder refines these cues: its depth-wise residual context enhancer reconstructs thin structures, and an attention gate filters out background noise. The result is a noise-robust, edge-accentuated feature map ready for final decoding.
\begin{figure*}[ht]
  \centering
  \begin{minipage}{1.0\linewidth}
  \centering
  \includegraphics[height=10cm]{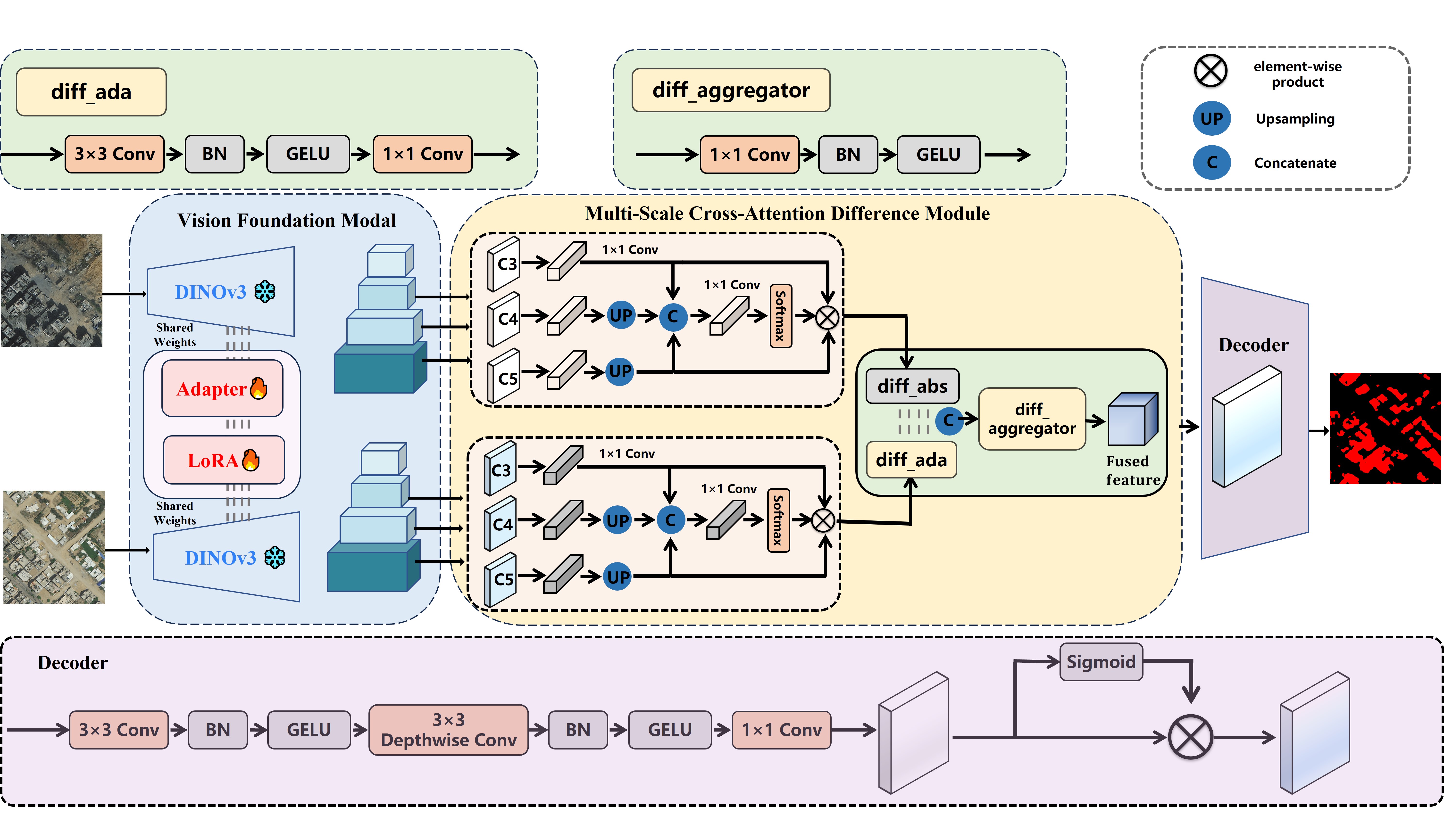}
  \end{minipage}\\[8pt]
  \caption{Overall architecture of the proposed MC-DiSNet.}
  \label{overall}
\end{figure*}

\subsection{Enhanced DINOv3 Backbone}

To maximize the feature extraction capability of the DINOv3 backbone while balancing computational cost and accuracy for the multi-class change detection (MCD) task, we introduce an enhanced DINOv3 encoder. This encoder keeps the original ConvNeXt-Tiny weights frozen and enriches multi-scale features C\textsubscript{3}-C\textsubscript{5} through three lightweight plug-ins. First, a bottleneck adapter is inserted after each MLP block, comprising a down-projection, GELU activation~\cite{lee2023mathematical}, and an up-projection branch with a reduction ratio. Its forward pass is formulated as:
\begin{equation}
\mathbf{y}=\mathbf{x}+s\cdot\mathrm{Up}\bigl(\mathrm{GELU}(\mathrm{Down}(\mathbf{x}))\bigr),
\end{equation}
where the scalar gate $s$ is initialized to zero and learned during training. Subsequently, 20 learnable prompt tokens are concatenated with the C\textsubscript{2}-C\textsubscript{4} feature maps and updated via cross-stage projection. This steers the network toward change-related semantics without modifying any backbone weights. Finally, low-rank adaptation (LoRA) is applied to both the attention projections and MLP layers in the last four blocks of the DINOv3 backbone, with a rank of 24, $\alpha$ of 48, and dropout of 0.1. This enables task-specific tuning with only 0.97M trainable parameters.

LoRA~\cite{hu2022lora} is a technique that maintains the 1.1 billion-parameter DINOv3 backbone frozen while inserting trainable rank-$r$ matrices into the last four blocks, specifically the attention projections and MLPs. This method allows for task-specific tuning without modifying the original backbone weights.

For any pre-trained weight $\mathbf{W}_0 \in \mathbb{R}^{d \times k}$, the forward pass is modified as follows:
\begin{equation}
\label{eq1}
    \mathbf{h} = \mathbf{W}_0 \mathbf{x} + \frac{\alpha}{r} \mathbf{B} \mathbf{A} \mathbf{x},
\end{equation}
where $\mathbf{B} \in \mathbb{R}^{d \times r}$ and $\mathbf{A} \in \mathbb{R}^{r \times k}$ are the trainable matrices, and only $\mathbf{B}$ and $\mathbf{A}$ are updated.

The rank $r$ controls the expressiveness of the model. A larger $r$ provides more basis vectors and finer corrections, but the parameter count grows approximately as $2dr$. The scaling factor $\alpha$ sets the effective step-size. An $\alpha \approx 2r$ keeps the initial update variance close to full fine-tuning. If $\alpha \ll r$, the adapter is suppressed, and if $\alpha \gg r$, it risks causing unstable gradients. 


\subsection{Multi-Scale Cross-Attention Difference Module}
To further enhance spatio-temporal information extraction from bi-temporal remote sensing images, we propose the multi-scale cross-attention difference (MSCAD) module as a core component of our framework. As illustrated in Figure~\ref{overall}, the MSCAD module proceeds in three conceptual steps: cross-scale alignment, temporal-difference modeling, and discrepancy aggregation.

First, multi-level features $C_{3}$, $C_{4}$, $C_{5}$ are projected to a common 256-D space via $1\times1$ conv, up-sampled to $C_{2}$ resolution, and fused by a channel-wise attention gate by
\begin{equation}
\mathbf{A} = \mathrm{Softmax}\bigl(\mathrm{Conv}_{1\times1}([\tilde{\mathbf{C}}_{3};\tilde{\mathbf{C}}_{4};\tilde{\mathbf{C}}_{5}])\bigr), 
\end{equation}
and
\begin{equation}
\mathbf{C}_{\mathrm{fused}} = \sum_{i=2}^{4} A_{i} \odot \tilde{\mathbf{C}}_{i}.
\end{equation}

Next, the concatenated tensor $\mathbf{F}_{\mathrm{cat}}$ $=$ $ [\mathbf{C}_{\mathrm{fused}}^{t_1};\mathbf{C}_{\mathrm{fused}}^{t_2}]$ is fed to a lightweight stack of $3\times3$ conv, BN, GELU and $1\times1$ conv to model non-linear temporal discrepancies and produce an adaptive difference feature $\mathbf{D}_{\mathrm{ada}}$ (diff\_ada in Figure \ref{overall}). 

Finally, the direct difference $\mathbf{D}_{\mathrm{dir}}=|\mathbf{C}^{t_1}-\mathbf{C}^{t_2}|$ (diff\_abs in Figure \ref{overall}) and the adaptive difference $\mathbf{D}_{\mathrm{ada}}$ are concatenated and compressed by $1\times1$ conv, BN and GELU, yielding the multi-scale change feature diff\_aggergator: 
\begin{equation}
\mathbf{D}_{\mathrm{out}} = \mathrm{GELU}\bigl(\mathrm{BN}\bigl(\mathrm{conv}_{1\times1}([\mathbf{D}_{\mathrm{dir}};\mathbf{D}_{\mathrm{ada}}])\bigr)\bigr).
\end{equation}

\subsection{Decoder}
To refine the coarse change cues produced by the MSCAD module, we design a enhancement decoder, which contains two lightweight yet effective sub-blocks.  First, a three-layer depthwise-separable residual stack (Figure~\ref{overall}) enriches spatial context via
\begin{equation}
\mathbf{Y} = \mathbf{X} + \mathrm{BN}\bigl(\mathrm{GELU}(\mathrm{DWConv}_{3\times3}(\mathrm{Conv}_{3\times3}(\mathbf{X})))\bigr),
\end{equation}
where $\quad \mathbf{X}\in\mathbb{R}^{C\times H\times W}$, followed by a $1\times1$ projection that recalibrates channel statistics, sharpening thin change edges and recovering detail lost during up-sampling.  Next, an element-wise sigmoid gate re-weights the refined difference map as
\begin{equation}
\mathbf{G} = \sigma(\mathbf{Y}),
\end{equation}
and we have:

\begin{equation}
     \mathbf{F}_{\mathrm{out}} = \mathbf{G}\odot\mathbf{Y},
\end{equation}
suppressing background noise and highlighting small or low-contrast changes. 
The output $\mathbf{F}_{\mathrm{out}}$ is subjected to a 4$\times$ upsampling operation to yield noise-robust, edge-sharp change predictions.

\subsection{Loss Function and Optimisation}

To effectively train the proposed multi-scale cross-attention difference siamese networ (MC-DiSNet), We minimise a composite loss
\begin{equation}
\mathcal{L}=0.4\,\mathcal{L}_{\text{Focal}}+0.3\,\mathcal{L}_{\text{Dice}}+0.3\,\mathcal{L}_{\text{Lovász}},
\end{equation}
whose three constituents are detailed below.

\textbf{Focal loss.}~\cite{lin2017focal} counter-acts foreground/background imbalance by down-weighting easy pixels and up-weighting hard ones. With class-balanced weight~$\alpha$ and focusing factor~$\gamma=3$,
\begin{equation}
\mathcal{L}_{\text{Focal}}=-\alpha(1-p_t)^{\gamma}\log p_t,
\end{equation}
where $p_t$ denotes the predicted probability of the ground-truth class.

\textbf{Dice loss.}~\cite{li2020dice} Dice loss maximises the overlap between prediction and ground-truth masks. Let $y\in\{0,1\}$ and $\hat{y}\in[0,1]$ be the ground-truth and predicted probabilities, respectively, then
\begin{equation}
\mathcal{L}_{\text{Dice}}=1-\frac{2\sum y\hat{y}+\varepsilon}{\sum y^{2}+\sum\hat{y}^{2}+\varepsilon},
\end{equation}
with $\varepsilon=10^{-5}$ for numerical stability.

\textbf{Lovász-Softmax loss.}~\cite{berman2018lovasz} Lovász-Softmax loss is a convex surrogate of the IoU that directly optimises the intersection-over-union measure. For a mini-batch of $N$ pixels and $C$ classes,
\begin{equation}
\mathcal{L}_{\text{Lovász}}=\sum_{c=1}^{C}\frac{1}{|\mathcal{Y}_c|}\sum_{i\in\mathcal{Y}_c}\Delta_{\text{IoU}}(p_{ic}),
\end{equation}
where $\mathcal{Y}_c$ indexes pixels of class~$c$ and $\Delta_{\text{IoU}}$ is the Lovász extension of the IoU loss.


\section{Experiments}

\subsection{Experimental Setup}

\subsubsection{Datasets Description}

In this section, we describe all the datasets we used, the evaluation metrics, and the experimental settings. There are three datasets to evaluate the proposed change detection method, one of which is a new dataset that we provide.

\begin{figure*}[ht]
  \centering
  \includegraphics[height=8cm]{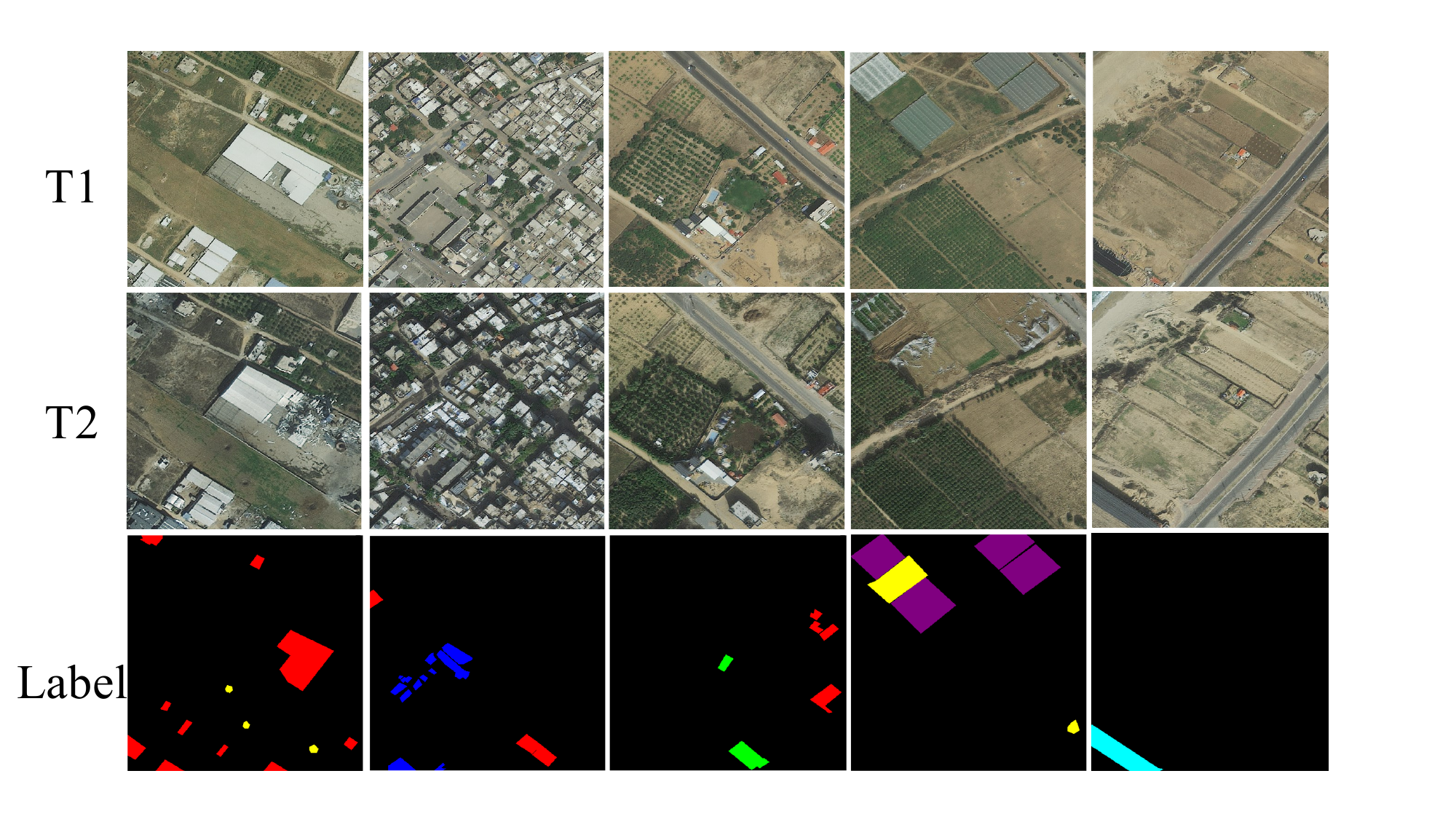}

  \caption{Examples of the proposed Gaza-Change, six distinct colors to encode change categories: red highlights “Building Damage", green marks “New Building", blue indicates “New Camp", yellow denotes "Farmland Damage", purple signals "Greenhouse Damage", whereas cyan represents "New Greenhouse".}
  \label{gazadata}
\end{figure*}

We release the Gaza-Change dataset, the first open-source bi-temporal benchmark for the Gaza Strip ($31.4^\circ N$-$31.6^\circ N$, $34.4^\circ N$-$34.6^\circ N$). It contains 922 precisely co-registered image pairs $512\times512$ acquired by the Beijing-2 satellite ($3.2m$ GSD) during 2023-2024. The imagery covers nine major urban areas, including Khan Yunis and Rafah. In line with our novel paradigm that focuses annotations exclusively on altered areas rather than full bi-temporal semantics, each pixel is meticulously annotated with one of six fine-grained change categories: \textit{building damage}, \textit{new building}, \textit{new camp}, \textit{farmland damage}, \textit{greenhouse damage}, and \textit{new greenhouse}. The dataset is randomly split into 554 training, 184 validation, and 184 test pairs. As shown in Figure~\ref{gazadata}, $T1$ and $T2$ show the bi-temporal images, and the label image illustrates the change labels with different colors. Beyond filling the data gap for semantic change detection task in conflict zones, Gaza-Change facilitates multi-dimensional humanitarian assessment and promotes scenario-specific algorithm development. Its six-class taxonomy characterizes the nature of each transition rather than a simple binary change, offering richer evidence for policy-making and resource allocation.

The SEmantic Change detectiON Dataset (SECOND) is a large-scale, pixel-wise annotated benchmark designed for SCD in high-resolution aerial imagery \cite{yang2021asymmetric}. It consists of 4662 bi-temporal image pairs ($512 \times 512$ pixels), collected over the Chinese cities of Hangzhou, Chengdu, and Shanghai using multiple platforms and sensors. Each pair was independently labeled by remote sensing experts to ensure high annotation fidelity. In contrast to our targeted annotation paradigm, SECOND provides exhaustive semantic labels for both time points, distinguishing six fundamental land-cover classes: \textit{non-vegetated ground surface, tree, low vegetation, water, buildings}, and \textit{playgrounds.}

Landsat-SCD~\cite{yuan2022transformer} provides ten change types and is constructed based on Landsat-like images acquired between 1990 and 2020 in Tumushuke, Xinjiang. The region is located along the Belt and Road Economic Belt, adjacent to the Taklimakan Desert, and features a fragile ecological environment. The data source consists of Landsat series imagery with a spatial resolution of 30 meters, which offers relatively high spatiotemporal and spectral resolution as well as good data accessibility.


Table \ref{tab:dataset} summarizes the specific information of the three aforementioned datasets. It is important to emphasize that while both the Gaza-Change, SECOND, and Landsat-SCD datasets are annotated with semantic labels. To validate the MCD task, the labels in the SECOND and Landsat-SCD dataset were also processed to retain semantic information exclusively for the changed areas.

\begin{table*}[h]
\centering

\caption{Main information of three datasets.}
\label{tab:dataset}
\begin{tabular}{ccccccc}
\toprule
Dataset & Patch Size & Resolution & Types of Changes & Train & Val & Test \\
\toprule

Gaza-Change & 512$\times$512 & 3.2m & 6 & 554 & 184 & 184 \\
SECOND & 512 $\times$512 & 0.5-3m & 30 & 2968& - & 1694 \\
Landsat-SCD & 416 $\times$ 416 & 30m & 9 &  1431&477&477 \\
\bottomrule
\end{tabular}
\end{table*}

\subsubsection{Evaluation Metrics}

In the domain of change detection, several traditional metrics are commonly used to evaluate the performance of models. These include overall accuracy (OA), precision (P), recall (R), mean intersection-over-union (mIoU), and F1 score. Each metric provides a different perspective on the model's performance, and they are crucial for understanding the effectiveness of change detection algorithms. Mathematically, let \( \mathrm{TP}_{c} \) denote the number of true positives for class \( c \), and \( \mathrm{FP}_{c} \) denote the number of false positives for class \( c \). Then \( C \) is the total number of classes, the key metrics are as follows:

\begin{gather}
   \mathrm{OA} = \frac{\sum_{i=0}^{N} q_{ii}}{\sum_{i=0}^{N}\sum_{j=0}^{N} q_{ij}} \\
   \mathrm{Precision}_{c} = \frac{\mathrm{TP}_{c}}{\mathrm{TP}_{c} + \mathrm{FP}_{c}}\\
   \mathrm{Recall}_{c} = \frac{\mathrm{TP}_{c}}{\mathrm{TP}_{c} + \mathrm{FN}_{c}} \\
   \mathrm{mIoU} = \frac{1}{C} \sum_{c=1}^{C} \frac{\mathrm{TP}_{c}}{\mathrm{TP}_{c} + \mathrm{FP}_{c} + \mathrm{FN}_{c}} \\
   \mathrm{F1}_{c} = 2 \cdot \frac{\mathrm{Precision}_{c} \cdot \mathrm{Recall}_{c}}{\mathrm{Precision}_{c} + \mathrm{Recall}_{c}},
\end{gather}
where \( q_{ii} \) represents the number of correctly classified pixels for class \( i \), and \( q_{ij} \) represents the number of pixels predicted as class \( i \) but actually belonging to class \( j \).

\subsubsection{Experimental Details}

All experiments were conducted on a single Nvidia RTX3090Ti GPU with 24 GB VRAM. Every model was trained for 200 epochs with a batch size of 4. A common augmentation pipeline (random horizontal, vertical flip, and random rotation) was applied to all training samples. All compared methods and our proposal share the identical optimizer recipe: AdamW~\cite{wen2025fantastic} with learning rate $3\times10^{-4}$, weight decay $0.01$, and $\beta=(0.9, 0.999)$. The learning rate is scheduled by cosine annealing with warm restarts ($T_0=30$, $T_{\text{mult}}=2$, $\eta_{\min}=1\times10^{-7}$). 

To accelerate the learning of newly introduced lightweight modules, we apply parameter-wise LR scaling (ours only) by multiplying the base learning rate as follows: \texttt{backbone.adapters} and \texttt{backbone.prompt\_tokens} by 20, \texttt{decoderhead} by 8, and \texttt{backbone} (frozen) by 0. All implementations are built on PyTorch~2.1.1 and trained from scratch under the above unified configuration.

All compared methods are described as follows:

\begin{itemize}

\item SNUNet~\cite{fang2021snunet}. Employs densely-connected nested UNet architecture with channel attention modules to enhance feature reuse efficiency and improve detail preservation in change detection tasks.
\item LGPNet~\cite{liu2021building}. Utilizes local-global pyramid networks to capture multi-scale building features, combined with attention mechanisms to enhance boundary accuracy in urban building change detection.
\item BIT~\cite{chen2021remote}.  A transformer-based framework for bi-temporal image processing, using semantic tokens to efficiently model change regions in remote sensing imagery
\item ChangeFormer~\cite{bandara2022transformer}. A pure transformer architecture for change detection that extracts multi-scale global features through hierarchical feature pyramids, significantly improving long-range dependency modeling. 
\item SARASNet~\cite{chen2023saras}. Incorporate relation-aware, scale-aware, and interaction modules to enhance the spatial and scale perception capabilities of the Siamese network.
\item STNet~\cite{ma2023stnet}. A remote sensing change detection network that introduces cross-temporal gating and cross-scale attention mechanisms for spatiotemporal feature fusion
\item USSFCNet~\cite{10081023}. Unified spatial-spectral frequency channel network that utilizes frequency domain transformations to mine deep feature representations, improving change detection robustness in complex scenarios.
\item DDLNeT~\cite{ma2024ddlnet}. A dual-domain learning network for remote sensing change detection that enhances change features in the frequency domain using discrete cosine transform while recovering spatial details.
\item ChangeMamba~\cite{chen2024changemamba}. Lightweight architecture based on state space models, achieving efficient long-sequence modeling through selective scanning mechanisms while balancing global receptive fields and computational efficiency. 
\item Rsmamba~\cite{liu2024rscama}. Visual state space model designed for remote sensing images, combining convolutional locality advantages with state space sequence modeling capabilities to optimize global context capture.
\item CDxLstm~\cite{wu2025cdxlstm}. Hybrid architecture integrating dilated convolutions with LSTM, enhancing detection performance for gradual changes through multi-scale temporal feature extraction.

\end{itemize}

\subsection{Main Results}

\textbf{Result of Caza-change dataset}. We report the results achieved by the model checkpoint that performs best on the test set. Table \ref{tab:complexity} summarizes the quantitative comparison between MC-DiSNet and nine representative methods.

\begin{figure*}[ht]
  \centering
  \begin{minipage}{1.0\linewidth}
  \centering
  \includegraphics[height=8cm]{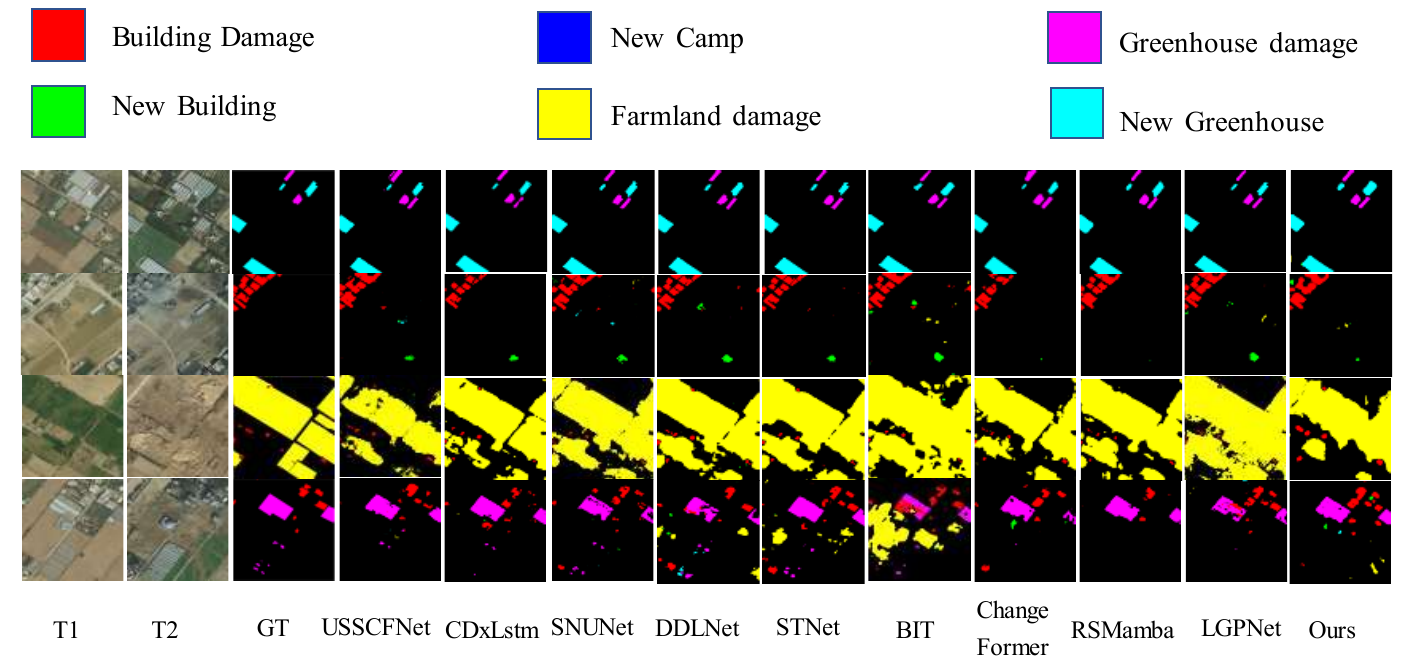}
  \end{minipage}\\[8pt]
  \caption{Example results on Gaza-change dataset.}
  \label{result1}
\end{figure*}


As reported in Table \ref{tab:complexity}, our MC-DiSNet achieves 86.10\% precision, 60.98\% recall, 55.16\% mIoU, and 69.25\% F1 score. The proposed MC-DiSNet exhibits a highly lightweight architecture, containing only 0.97 million parameters, which is 42 times fewer than ChangeFormer and 12 times fewer than BIT. Our framework outperforms the previous best method, DDLNet, by a large margin of 12.0\% in F1 score and 12.9\% in mIoU, underscoring its significant advantages in accuracy and efficiency.

\begin{figure*}[ht]
  \centering
  \includegraphics[height=6.5cm]{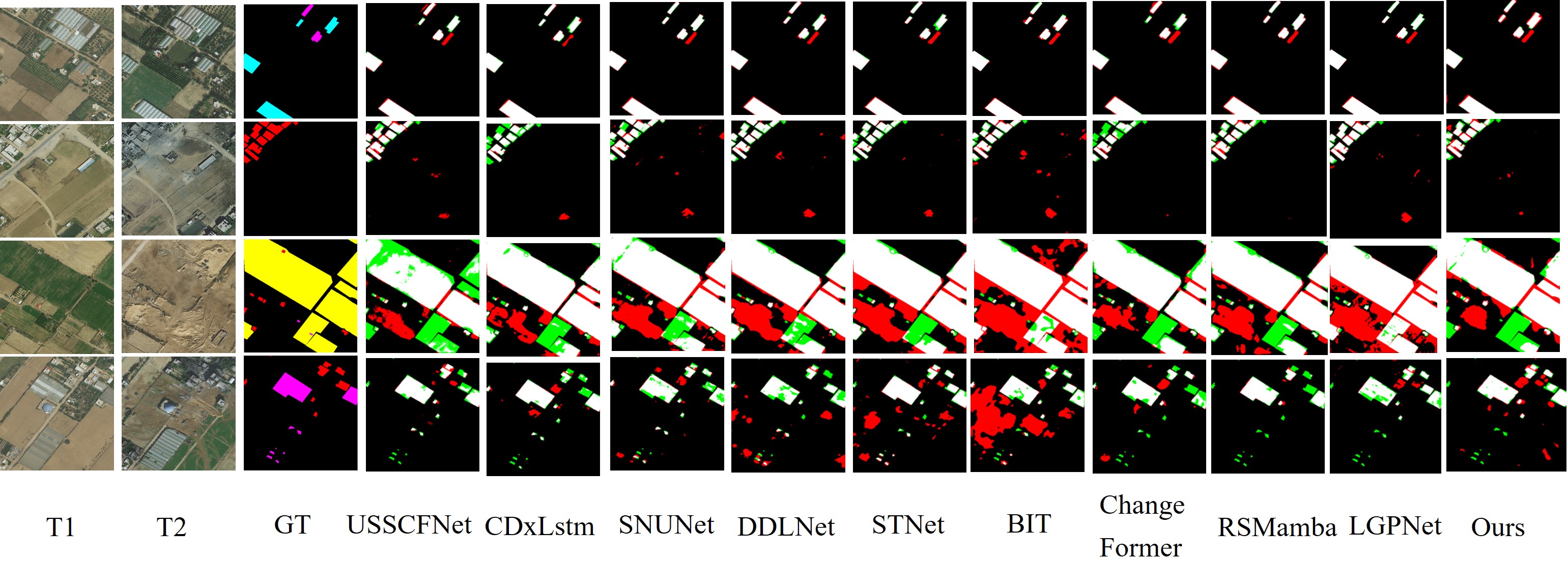}

  \caption{Qualitative comparison on the Gaza-Change dataset, with color coding: true positives (white), true negatives (black), false positives (red), and false negatives (green).}
  \label{compare}
\end{figure*}

As shown in Figure~\ref{result1} and Figure~\ref{compare}, we provide representative visual comparisons of different change detection methods on sample cases from the Gaza-Change dataset. These qualitative results clearly demonstrate our method's advantages. As shown in the first row, for scenarios with relatively distinct categories and simpler contexts, most methods can accurately identify semantic changes to a certain extent. The second row demonstrates that for relatively dense building damage, methods like CDxLstm and ChangeFormer exhibit fewer false positives but suffer from more missed detections. In comparison, our method achieves a better balance between overall false positives and false negatives. From the third and fourth rows, it can be observed that for agricultural areas and weed fields with highly similar shapes and colors, most models struggle to achieve accurate identification. Traditional convolutional methods such as USSCFNet, SNUNet, and LGPNet tend to produce discontinuous patchy detection results, whereas our method generates the fewest false alarms while maintaining detection continuity.

\begin{table*}[!t]
\centering
\caption{Comparison with mainstream change detection methods on the Gaza-change dataset.}
\label{tab:complexity}
\begin{tabular}{ccccccc}

\toprule 
Method & {Trainable Params (M)} & P(\%)  & R(\%)  & mIoU(\%)  &{F1}(\%) & Best Epoch \\
\hline
USSFCNET     & 1.5  &  46.66  & 48.75 & 34.43  &47.20 &  51 \\ 
BIT          & 11.9 & 57.45 & 54.90 & 42.13 &56.15 &  65  \\
SNUNet       & 12.0 & 51.44 &51.42 &36.50 & 50.35& 49   \\
DDLNET       & 13.8 & 56.61&59.96 & 42.23 &57.22 & 34 \\
STNet        & 14.6 & 59.15 &54.96 &42.81 & 56.64& 53 \\
CDxLstm      & 16.2 & 56.96 & 47.29 &36.84 & 51.25& 83 \\
ChangeFormer & 41.1 & 59.06 & 46.52& 36.62  &50.50 & 88 \\
Rsmamba      & 52.0 & 58.43 & 42.81 & 36.45 &49.40 & 66 \\
LGPNet       & 71.0 & 55.69 & 55.69& 48.64 &  54.64&  64 \\ 
Ours   & 18.2  & \textbf{86.10} & \textbf{60.98}  & \textbf{55.16} &\textbf{69.25} & \textbf{29} \\

\bottomrule
\end{tabular}
\end{table*}

\textbf{Results of SECOND and Landsat-SCD dataset}. To further verify the effectiveness and generalizability of the proposed method, we conduct experiments on the two extra SCD datasets: SECOND and Landsat-SCD. Quantitative results are summarised in Table~\ref{tab:second}. 

Obviously, our method achieves an mIoU of 27.74\% and an F1-score of 40.59\%, surpassing the strongest competitor (BIT) by 6.87 percentage points in mIoU and 8.39 percentage points in F1-score.
On the large-scale Landsat-SCD dataset, our method achieves a mIoU of 68.79\% and an F1-score of 80.69\%, significantly outperforming the previous state-of-the-art LGPNet (61.48\% mIoU and 75.45\% F1). This represents notable absolute gains of +7.31\% in mIoU and +5.24\% in F1, corresponding to relative improvements of 11.9\% and 6.9\%, respectively. This consistent superiority on two external datasets demonstrates that the proposed modules do not overfit the primary training data and generalize effectively to other remote sensing scenarios with varying imaging conditions and annotation granularity.

\begin{table*}[ht]
  \centering
  \caption{Performance comparison on SECOND and  Landsat-SCD datasets.}
  \label{tab:second}
  \begin{tabular}{lccccc}
    \toprule
    \multirow{2}{*}{Method} & \multicolumn{2}{c}{SECOND} & \multicolumn{2}{c}{ Landsat-SCD} \\
    \cmidrule(lr){2-3} \cmidrule(lr){4-5}
     & mIoU (\%) & F1 (\%) & mIoU (\%) & F1 (\%) \\
    \midrule
    BIT \cite{}          & 20.87 & 32.20 &60.13  & 74.33 \\
    SNUNet \cite{}       & 12.88 & 20.85 &42.07  & 56.28 \\
    USSFCNet \cite{}     & 12.03 & 17.80 & 41.05 & 55.14  \\
    ChangeFormer \cite{} &15.48  &28.30  & 57.09 & 71.45 \\
    CDxLstm \cite{}      & 15.19 & 27.25 & 47.93& 62.89 \\
    RSMamba \cite{}      & 12.54 & 20.09 &55.41 &70.00  \\
    LGPNet \cite{}      & 12.42 & 19.37 &61.48 &75.45  \\
    Ours        & \textbf{27.74} & \textbf{40.59} & \textbf{68.79} & \textbf{80.69} \\
    \bottomrule
  \end{tabular}
\end{table*}

\begin{figure*}[ht]
  \centering
  \begin{minipage}{1.0\linewidth}
  \centering
  \includegraphics[height=7.5cm]{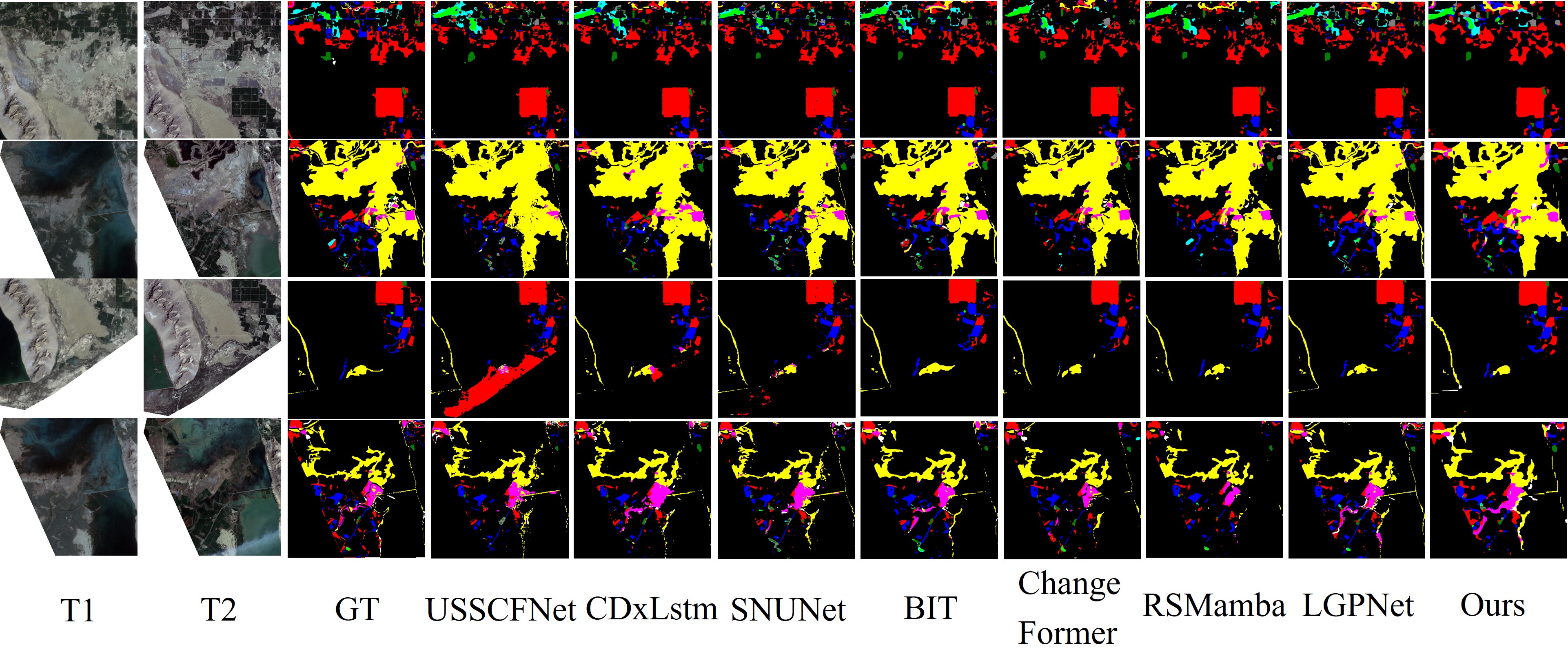}
  \end{minipage}\\[8pt]
  \caption{Example results on Landsat-SCD dataset.}
  \label{landsat1}
\end{figure*}

\begin{figure*}[ht]
  \centering
  \includegraphics[height=6.5cm]{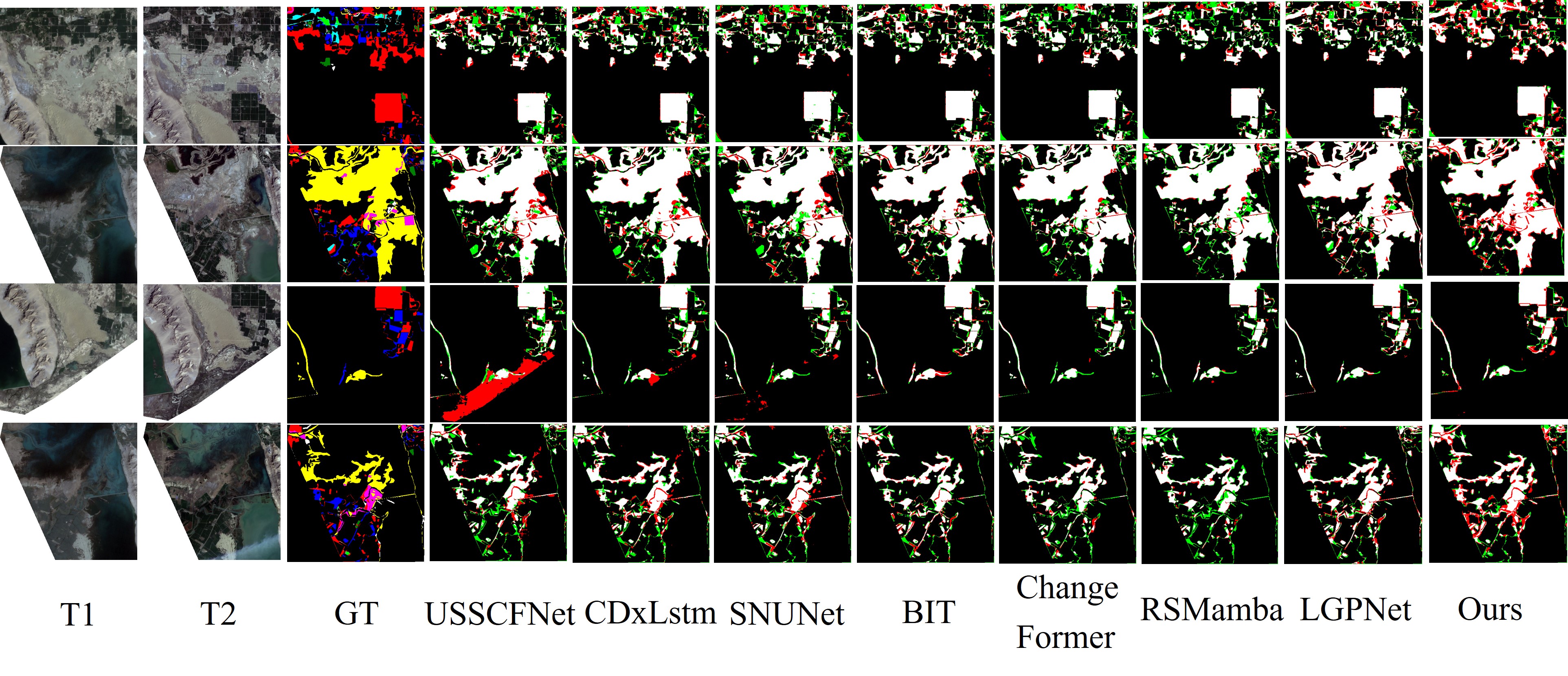}

  \caption{Qualitative comparison on the Landsat-SCD dataset, with color coding: true positives (white), true negatives (black), false positives (red), and false negatives (green).}
  \label{landsat2}
\end{figure*}


We also present visual detection results on the Landsat-SCD dataset in Figure~\ref{landsat1} and Figure~\ref{landsat2}. As shown in the third row of Figure~\ref{landsat1}, MC-DiSNet produces a continuous and uniformly wide delineation of the narrow, curved change belt, whereas results from other methods exhibit fragmentation or over-expansion. The first row of Figure~\ref{landsat2} demonstrates that our result aligns closely with the ground truth (GT) without spurious speckles, while other methods show extensive green false negatives (FN). However, it is important to note that while our method significantly reduces under-detection, it introduces a certain amount of false positives (red pixels), leading to an overall reddish hue in some areas. This observation underscores a broader challenge: effective change detection models often require targeted adaptation to the specific characteristics of different remote sensing datasets.

Overall, As demonstrated on our own dataset as well as on the additional benchmarks, the proposed approach consistently delivers state-of-the-art results not only on small-scale sets (hundreds of image pairs) but also on large-scale collections (tens of thousands of samples). This robust performance across dramatically different data volumes indicates strong generalization capability and practical deployment potential for diverse remote-sensing scenarios.

\begin{table*}[ht]
  \centering
  \caption{Module-level ablation results on the validation set. Best scores are in \textbf{bold}.}
  \label{tab:module_ablation}
  \begin{tabular}{lclclclclclclcl}
\toprule
  Module & MS-att & diff-ada & diff-agg & att  & P(\%) & R(\%) & mIoU (\%) & F1 (\%) \\
\hline
  
   \multirow{3}{*}{Encoder}&  &\checkmark& \checkmark &\checkmark  &\textbf{89.37} & 54.35 &  50.28& 64.99 \\
&\checkmark & & \checkmark& \checkmark  & 79.26 & 49.20 & 44.25 &   58.36 \\
 & \checkmark & \checkmark& & \checkmark& 78.77 & 50.50 & 46.10& 59.09\\

  Decoder  &\checkmark & \checkmark &\checkmark &&  83.39 & 60.24 & 53.81& 67.26  \\
   Base & \checkmark & \checkmark & \checkmark& \checkmark&  86.10 & \textbf{60.98} & \textbf{55.16}& \textbf{69.25} \\
\bottomrule
  \end{tabular}
\end{table*}

\begin{table*}[!t]
\centering
\caption{Comparison with different backbones in encoder of our proposed MC-DiSNet on the Gaza-change dataset.}
\label{tab:backboneablation}
\begin{tabular}{ccccccc}

\toprule 
Method &P(\%)  & R(\%)  &  mIoU(\%)  &{F1}(\%) & {Params (M)} & Trainable Params (M) \\
\hline
Dinov3-tiny (ours)     & 86.10 &  60.98  & 55.16 & 69.25  & 29.09  &  1.78 \\ 
Dinov3-smal1          & 77.52 & 65.35 & 62.03 & 70.71 &53.06 &  3.25  \\
ResNet-18       & 51.56 & 40.13 &32.45 &42.99 & 14.95& 0.68  \\
ResNet-50       & 59.37 & 39.07&34.03 & 44.48 &25.76 & 1.55 \\
SAM-base        & 90.61 & 59.95 &56.46 &71.65& 95.39& 1.36 \\
Swinformer-tiny      & 73.71 & 63.47 & 58.05 &67.92 & 29.58& 1.62 \\

\bottomrule
\end{tabular}
\end{table*}

\begin{table}[h]
\centering
\caption{Ablation on LoRA rank $r$ and scaling factor $\alpha$.}
\label{tab:lora_hyper}
\begin{tabular}{@{}ccccc@{}}
\toprule
$r$ & $\alpha$ & Params(M) & mIoU (\%) & F1 (\%) \\
\hline

4  & 4  & 0.16 &  54.30 & 68.30   \\
4  & 8  & 0.16 &  51.70 & 66.21  \\
8  & 8  & 0.32 & 52.75 & 67.43    \\
8  & 16  & 0.32 & 51.08 & 65.47  \\
16  & 16 &0.64 & 52.68  &66.08    \\
16  & 32 & 0.64 & 53.81  & 68.25 \\
24  & 24 & 0.97 & 55.08 & 68.96 \\
\textbf{24} & \textbf{48} & \textbf{0.97}& \textbf{55.16} & \textbf{69.25}  \\
32 & 32 & 1.29 & 54.27& 68.89  \\
32 & 64 & 1.29 & 52.83 & 66.94  \\
\bottomrule

\end{tabular}
\end{table}

\begin{table*}[htpb]
\centering
\caption{Quantitative results of our MC-DiSNet on the proposed Gaza-change dataset under the MCD task, reporting per-category performance metrics for all six change types.}
\label{tab:xilidu}
\begin{tabular}{ccccccc}

\toprule 
Class & OA(\%) & P(\%)  & R(\%)  & F1(\%)  &IoU(\%)  \\
\hline
building damage    & 95.55  &  98.92  & 75.26 & 85.48  & 74.64  \\ 
new building      & 99.52 & 73.67 & 59.32 & 65.72 & 48.94  \\
new camp     & 99.57 & 70.21 &  44.10  &54.17& 37.15\\
farmland damage     & 97.69 & 99.00 & 28.95& 44.79& 28.86\\
greenhouse damage        & 99.74 & 89.24 & 69.98& 78.45& 64.54\\
new greenhouse     & 99.86 & 85.58 & 88.28     & 86.91& 76.85\\

\bottomrule
\end{tabular}
\end{table*}

\subsection{Ablation Study}

To systematically evaluate the contributions of each proposed component, we conduct two groups of ablation experiments: (1) module-level ablation on the multi-scale attention mechanism, and (2) parameter-level ablation on the LoRA-specific hyperparameters.

\subsubsection{Module-level Ablation}

We conduct ablation studies on key components of MC-DiSNet: (a) the multi-scale attention (MS-att) in the encoder, (b) the two feature fusion modules, diff\_Ada and diff\_agg, and (c) the attention mechanism (Att) in the decoder. Table \ref{tab:module_ablation} reports the comparison results.

To systematically validate the necessity and synergistic effects of the core components in MC-DiSNet, we conducted module ablation experiments on the validation set, with results presented in Table 4. The full model with all components (Base) achieved the best performance (55.16\% mIoU, 69.25\% F1). Removing components individually led to performance degradation to varying degrees: Removing the multi-scale attention (MS-att) caused mIoU to drop by 4.88\% and F1 by 4.26\%, indicating its crucial role in multi-scale discriminative feature selection. Removing the decoder attention (att) reduced mIoU to 53.81\% and lowered the recall rate, confirming its core function in refining the final change map. Removing diff\_agg resulted in one of the most severe performance drops (mIoU 46.10\%, F1 59.09\%), highlighting the irreplaceability of cross-level fusion of difference features. Finally, completely removing diff\_ada led to the largest degradation (mIoU 44.25\%, F1 58.36\%), demonstrating its key role in suppressing false-change noise.

To evaluate the robustness and selection dependency of MC-DiSNet on encoder backbones, we systematically compare six mainstream/lightweight backbone networks on the Gaza-change dataset, with results summarized in Table \ref{tab:backboneablation}. The experiments demonstrate that Dinov3-tiny achieves the optimal balance among accuracy, efficiency, and parameter count. Employed as a frozen self-supervised vision backbone, it requires only 1.78M trainable parameters (29.09M total) to attain 55.16\% mIoU and 69.25\% F1-score, while also achieving the fastest convergence and highest precision (86.10\%).  In contrast, Dinov3-small, although capable of further improving performance (62.03\% mIoU), doubles the trainable parameters (3.25M) and significantly increases the overall model size (53.06M), shifting the trade-off towards a heavier model. Traditional CNN backbones (ResNet-18/50) perform markedly worse (mIoU $\leq$ 34.03\%, F1 score $\leq$ 44.48\%), revealing the limitations of convolutional stacking in capturing long-range dependencies in remote sensing time series. SAM-base achieves the highest F1-score (71.65\%) and Precision (90.61\%) due to its large-scale mask pre-training, but its substantial parameter count of 95.39M (approximately 3.3$\times$ that of Dinov3-tiny) indicates significant redundancy. Swinformer-tiny delivers performance (58.05\% mIoU) between the two Dinov3 variants, but its hierarchical windowed attention mechanism introduces additional memory and computational overhead. In summary, Dinov3-tiny achieves the optimal balance among accuracy, parameter count, and training efficiency, making it the backbone of choice for MC-DiSNet in change detection tasks.

\subsubsection{Parameter-level Ablation }

Table~\ref{tab:lora_hyper} compares different combinations of rank $r$ and scaling factor $\alpha$. As shown in Table.\ref{tab:lora_hyper}, the best trade-off between accuracy and parameter budget is achieved at r=24 and $\alpha$=48, where the model reaches 69.25\% F1 and 55.16\% mIoU with only 0.97 M additional trainable weights.

Increasing r from 8 to 24 yields a +3.78\% F1 improvement, while further growing r to 32 drops F1 by 2.31\% and almost doubles the parameters, revealing a clear saturation point.

Maintaining $\alpha/r \approx 2$ consistently outperforms $\alpha/r=1$, confirming that a slightly larger scaling factor helps the low-rank adapter capture stronger change representations. Consequently, we adopt \textbf{$lora_r=24$} and \textbf{$lora{\alpha}=48$} as the default LoRA configuration in all subsequent experiments.

\section{Discussion}

Our study makes three primary contributions to conflict zone monitoring through remote sensing. First, we introduce a novel dataset specifically designed for assessing land cover changes in conflict-affected areas, categorized into six critical classes: building damage, new building, new camp, farmland damage, greenhouse damage, and new greenhouse. This dataset addresses a significant gap in available benchmarks for humanitarian damage assessment. 

Second, we introduce the MCD task that directly annotates changed areas rather than separately labeling bi-temporal semantic regions. This approach effectively extends traditional Binary Change Detection (BCD) to semantic change analysis while avoiding error accumulation from intermediate processing stages. The MCD framework substantially reduces both annotation workload and complexity, enabling rapid damage assessment in conflict scenarios. 

Third, to address the challenge of limited feature representation in small targets, we develop the Multi-scale Cross-attention Difference Siamese Network (MC-DiSNet), which leverages pre-trained foundation models to enhance feature extraction capability. As reported in Table \ref{tab:xilidu}, our method achieves promising performance on several critical damage categories, with F1-scores of 74.64\% for greenhouse damage, 64.54\% for new greenhouse, and 76.85\% for building damage.




The moderate performance on new building and new camp (45.21\% and 52.67\% respectively) likely stems from their relatively small spatial extent in the imagery. Notably, farmland damage shows the lowest performance (28.86\%), potentially due to the significant feature distribution differences between agricultural areas and other structural damage categories, suggesting that our model may require further adaptation to effectively capture diverse land cover characteristics. 

These findings highlight both the promise and challenges of automated damage assessment in conflict zones, while demonstrating the potential of the MCD paradigm and MC-DiSNet architecture to advance the field of humanitarian remote sensing.

\section{Conclusion}

In this work, we have introduced a comprehensive framework for semantic change detection in conflict zones through three key contributions. First, we presented a novel dataset specifically designed for assessing conflict-induced land changes, featuring six critical damage and construction categories. Second, we introduced the MCD paradigm, which directly focuses on changed areas to enable efficient damage assessment while significantly reducing annotation costs. Third, we developed the MC-DiSNet that effectively leverages pre-trained vision foundation models to enhance feature representation for small targets. Our experimental results demonstrate that the proposed method achieves strong performance on the Gaza-change, SECOND, and Landsat-SCD datasets. This work takes a significant step toward automated conflict damage assessment using remote sensing imagery. We believe our contributions in dataset creation, task formulation, and methodology development will facilitate future research in humanitarian remote sensing and emergency response.



\section*{Data availability}
The Gaza-change dataset presented in this study is available to qualified researchers upon reasonable request. 

\section*{Acknowledgments}
The authors extend sincere appreciation to the DeepSeek large language model for its valuable assistance in text polishing and language refinement during the writing of this paper.


\bibliographystyle{cas-model2-names}

\bibliography{cas-refs}





\end{document}